\documentclass[conference]{IEEEtran}
\IEEEoverridecommandlockouts
\usepackage{cite}
\usepackage{amsmath,amssymb,amsfonts}
\usepackage{algorithmic}
\usepackage{graphicx}
\usepackage{textcomp}
\usepackage{xcolor}

\def\BibTeX{{\rm B\kern-.05em{\sc i\kern-.025em b}\kern-.08em
    T\kern-.1667em\lower.7ex\hbox{E}\kern-.125emX}}
\begin{document}

\title{Trustworthy deep domain adaptation for wearable photoplethysmography signal analysis with decision-theoretic uncertainty quantification\\
\thanks{The project (22HLT01 QUMPHY) has received funding from the European Partnership on Metrology, co-financed from the European Union’s Horizon Europe Research and Innovation Programme and by the Participating States. Funding for NPL was provided by Innovate UK under
the Horizon Europe Guarantee Extension, grant numbers 10084125.}
}

\author{\IEEEauthorblockN{Ciaran Bench}
\IEEEauthorblockA{\textit{Department of Data Science and AI} \\
\textit{National Physical Laboratory}\\
Teddington, UK \\
ciaran.bench@npl.co.uk}

}

\maketitle

\begin{abstract}
In principle, deep generative models can be used to perform domain adaptation; i.e. align the input feature representations of test data with that of a separate discriminative model's training data. This can help improve the discriminative model's performance on the test data. However, generative models are prone to producing hallucinations and artefacts that may degrade the quality of generated data, and therefore, predictive performance when processed by the discriminative model. While uncertainty quantification can provide a means to assess the quality of adapted data, the standard framework for evaluating the quality of predicted uncertainties may not easily extend to generative models due to the common lack of ground truths (among other reasons). Even with ground truths, this evaluation is agnostic to how the generated outputs are used on the downstream task, limiting the extent to which the uncertainty reliability analysis provides insights about the utility of the uncertainties with respect to the intended use case of the adapted examples. Here, we describe how decision-theoretic uncertainty quantification can address these concerns and provide a convenient framework for evaluating the trustworthiness of generated outputs, in particular, for domain adaptation. We consider a case study in photoplethysmography time series denoising for Atrial Fibrillation classification. This formalises a well-known heuristic method of using a downstream classifier to assess the quality of generated outputs. 
\end{abstract}


\section{Introduction}
While deep learning models can be used to perform a wide range of tasks, the generalisation of their performance to unseen test data depends on the overlap in the training and test data domains. A data domain $\mathcal{D} = \{\chi, P(x), P(x,y)\}$ consists of an input feature space $\chi$ (a vector space that contains all input features), a marginal distribution $P(x)$, and a joint probability distribution $P(x,y)$. Here, $x\in X$ which is composed of $N$ training example inputs $x_1, x_2, ... x_N \in X$ and $y$ is a member of the corresponding set of ground truths $y_1, y_2, ...y_N \in Y$ \cite{kouw2019review}. 

In cases where there is a significant domain gap between $\mathcal{D}_{\text{train}}$ and $\mathcal{D}_{\text{test}}$, one can improve generalisability by performing domain adaptation, i.e. modifying the properties of the data to increase the overlap of their data domains \cite{farahani2020concise}. Among several possible approaches \cite{farahani2020concise}, we consider asymmetric test input domain adaptation, where we aim to align the input feature representations (and given our case study, also the joint distributions) of $\mathcal{D}_{\text{train}}$ and $\mathcal{D}_{\text{test}}$ by adapting the input feature representations of the test data. This is appealing given our case study involves the denoising of wearable photoplethysmography (PPG) signals for a pretrained Atrial Fibrillation (AF) classification model (Section \ref{sec:case}). Here, we ultimately aim to ensure the feature representations of the denoised signal and their relation to the classification target are similar to the classifier's training data. This can be achieved by simply modifying the input representations alone (we preserve target information, as this is derived from corresponding ECG measurements \cite{soto2020deepbeat}). We rule out the use of symmetric domain adaptation approaches given we use a classifier model pretrained on low noise PPG signals to perform the AF classification (therefore high noise examples that have been adapted to appear low noise are desired).

We consider the use of a 1D variant of a Pix2pix generative adversarial network (GAN) to perform the denoising task \cite{henry2021pix2pix}, given  generative deep learning models can achieve greater realism for time-series based adaptation tasks compared to more generic non-generative optimisation approaches \cite{brophy2022denoising,xu2021ecg}. However, generative models are prone to outputting hallucinations or other artefacts which can result in poor performance. We aim to use uncertainty quantification to assess the trustworthiness of the denoising model, and provide some indication as to whether the use of the generated output may result in improved classification performance when used as an input to the pretrained model.

Uncertainty quantification (UQ) is one of several possible strategies for evaluating the quality of outputs post-deployment \cite{fan2025trustworthiness}. This can help streamline the use of models by enabling users to disregard lower quality outputs. Compared to other quality indicators, UQ can provide an indication of trustworthiness in cases where there are no ground truths (unlike generic error measures like the mean squared error, or structural similarity index \cite{hore2010image}) or when there is a lack of a well formulated metric that can capture the characteristic feature representations of the data (e.g. the insensitivity of the Blind/Referenceless Image Spatial Quality Evaluator (BRISQUE) \cite{mittal2012no}). UQ can flag a single generation as uncertain (unlike global metrics like the Fr\'echet Inception Distance (FID) \cite{wu2025pragmatic}), and in principle can offer improved insight into possible causes of poor performance through the disentanglement of different sources of uncertainty (e.g. irreducible aleatoric uncertainty or reducible epistemic uncertainty \cite{kendall2017uncertainties}).

In typical measurement frameworks, a model is used to estimate the value of the quantity of interest, known as the measurand, where its uncertainty should reflect the doubt in the estimate \cite{iso1993guide}. It is desirable for this uncertainty to be calibrated, i.e. that the magnitude of the uncertainty estimate correlates with the magnitude of the prediction error \cite{pernot2023calibration,pernot2023can}. However, ground truths are not always available for generative deep learning tasks. This complicates the assessment of calibration, where instead per instance proxies of the negative log-likelihood (NLL) \cite{pang2023calibrating} or proxy metrics for accuracy like the FID are used instead \cite{bench2025trustworthy}. The use of the per-instance proxy described in \cite{pang2023calibrating} is not easily generalisable to other tasks/models as it relies on a specific mathematical
property of diffusion models. The FID, or analogous metrics that could be formulated for 1D time series, do not provide a good indication of individual/local reliability (i.e. reliability at the level of individual or few binned predictions), which is often desirable in practical applications where one or few examples are used to execute a given task \cite{pernot2023calibration, bench2026uncertainty}. Also, the use of proxy accuracy metrics like the FID (or variants that could in principle be developed for 1D time series), may not provide effective measures of prediction error for non-natural images. Even in cases where there are ground truths available (or where a proxy metric for accuracy may be suitable), this simple measurement framework considers whether the magnitude of the estimate of the measurand is accurate, and does not consider whether use of the uncertainty estimate in some decision-making process will result in the desired outcome. This is often what is of value in practical settings. This motivates the need for an uncertainty quantification framework that caters to the decision making process, and does not require some explicit notion of prediction error for the generated outputs to evaluate uncertainty reliability. We consider the decision-theoretic formulation described in \cite{smith2024rethinking}, as it is compatible with our use-case and it addresses other weaknesses with more standard UQ techniques.


\subsection{Case study: Denoising wearable PPG signals for Atrial Fibrillation classification}
\label{sec:case}
Blood flow encodes important physiological information about patient health. Photoplethysmography is an optical sensing modality where near infrared light is shone onto the skin (often at the vascular periphery, such as at the ear lobe or finger tip), where the variation in blood flow results in changes in the scattering and absorption coefficients of the tissue, and consequently, variation in the amount of transmitted or backscattered light collected by a sensor opposite or adjacent to the emission source. Time series indicating the variation in transmitted/backscattered light can therefore encode information about blood flow which relates to cardiac health. It has been show that PPG time series can be used to detect Atrial Fibrillation (short periods of rapid uncoordinated heartbeat) where variation in signal morphology encodes information about atypical heart rhythm \cite{soto2020deepbeat}. Deep learning models have been shown to be a particularly promising modelling approach, capable of handling the large amounts of data acquired from wearable sensors, and therefore, provide a potential means to continuously monitor the condition outside of the clinic \cite{aschbacher2020atrial,antiperovitch2024continuous}. 

We consider a deep learning model trained on data $\mathcal{D}_{\text{train}}$, composed of minimally pre-processed time series acquired from a custom split of the Deepbeat dataset \cite{soto2020deepbeat,bench2025uncertainty}, and corresponding binary classification labels for whether an AF episode occurred within the duration of the time series. The split ensures there is no patient overlap and a balanced label distribution across the training (106,249 examples), validation (15,256 examples), and test sets (15,377 examples). The data is composed of raw time-series samples with durations of 25 s (sampling rate 32 Hz) that have been preprocessed using low-pass, high-pass, and adaptive filters to remove artefacts, noise, and baseline wander. 
We use the pretrained model described in \cite{bench2025uncertainty} (a 1D variant of AlexNet trained with stochastic gradient descent and a loss given in \cite{bench2025uncertainty}). 

To construct our case study in PPG denoising, the test inputs $\mathcal{D}_{\text{test}}$ are augmented with random Gaussian noise where the augmented dataset is given as $\mathcal{D}^{*}_{\text{test}}$, so the domain gap between $\mathcal{D}_{\text{train}}$ and $\mathcal{D}^{*}_{\text{test}}$ is larger compared to $\mathcal{D}_{\text{train}}$ and $\mathcal{D}_{\text{test}}$. We then consider the use of deep domain adaptation to denoise the augmented test data, where decision-theoretic uncertainty quantification (DTUQ) is used to determine whether the generated output is likely to result in a correct prediction with the pretrained classifier. Given the aim here is mainly to illustrate the utility of the DTUQ approach, we only consider random additive Gaussian noise as opposed to more realistic noise models \cite{masinelli2021synthetic}.

\subsection{Decision-theoretic uncertainty quantification}
DTUQ provides a notion of uncertainty that is grounded in actions and the consequences of taking these actions\cite{smith2024rethinking}. Here, a utility function $l$ defines the consequences of taking  an action or decision $a\in A$ in order to realise some ground truth outcome $z \in Z$. $l$ is used to select the optimal action, where subjective beliefs over the possible outcome $z$, which are not known \textit{a priori}, are used to acquire an average expected value of $l$ given $a$ (i.e. a subjective expected loss). One chooses $a$ based on that which provides the lowest subjective expected loss. $l$ is formulated based on a subjective assessment of the decision cost.

For a predictive model $p_n(z) = p(z;y_{1:n})$ tasked with estimating realisations of $z$, where $y_{1:n}$ is the training data and $n$ is the amount of data, the action that minimises the subjective expected loss is Bayes optimal, where $p_n(z)$ can be interpreted as a Bayes estimator, i.e. the prediction of the model defines beliefs about $z$, so the Bayes optimal action is given by:
\begin{equation}
a^* = \arg\min_{a \in A} \, \mathbb{E}_{p_n(z)}\!\left[l(a,z)\right].
\end{equation}
For example, in the case where log loss is used as a decision cost, the expectation $E_{p_n(z)}$ is expressed as the entropy of predictions when the action is predicting the full predictive distribution, i.e. $H[p_n(z)] = E_{p_n}[l(a^*_n,z)]$. Generally speaking, if a loss function can be defined based on subjective preferences related to decision making, an expression of uncertainty naturally follows which reflects preferences about model behaviour with respect to the final decision made from the prediction.

An evaluation of the reliability of the predicted uncertainties should involve external grounding; i.e. do the use of uncertainties result in lower decision cost? In our case where the performance of an auxiliary classifier represents the decision cost (misclassification rate), the evaluation procedure involves comparing classification accuracy with the expression of uncertainty.

Here, we use a generative deep learning model to adapt PPG time series, where a decision is made about whether the use of the adapted example as an input to an auxiliary deep learning model trained to perform AF classification will likely result in a correct prediction. Observing features $x$ we aim to choose an action $a$ with an associated loss. The decision cost is the misclassification loss given by: 
\begin{equation}
    L(a,y) = 1\{a\neq y\}.
\end{equation} The conditional risk/posterior expected loss that we hope to minimise (loss weighted by the posterior probability of each outcome, summed over all outcomes which is the expected loss of predicting class $a$ given features $x$ under the posterior belief of possible true classes $y$), 
\begin{equation}
    \rho(a|x) = \sum_y L(a,y)p(y|x),
\end{equation}
\begin{equation}
   = \sum_y 1\{a\neq y\}p(y|x),
\end{equation}
\begin{equation}
   = \sum_{y\neq a}p(y|x).
\end{equation} Given $\sum_{y\in Y}p(y|x)=p(y=a|x) + \sum_{y\neq a}p(y|x)$, and $\sum_{y\in Y}p(y|x)=1$, the term 

\begin{equation}
    \sum_{y\neq a}p(y|x) = 1-p(a|x).
\end{equation}

Therefore 
\begin{equation}
    \rho(a|x) =  1-p(a|x),
    \label{eq:rho}
\end{equation} where we aim to choose an action that minimises this expression. This is equivalent to $1-\text{max}(p(y|x))$ \cite{chen2025efficient}.

We use the predictive entropy of the downstream classifier as a proxy for decision cost to make use of the Uncertainty Calibration Error - a known reliability metric (the entropy is monotonically linked to misclassification/the maximum of the predicted probability in binary classification, and commonly observed to correlate with misclassification in practice \cite{lavesuncertainty,bench2025uncertainty}). We externally ground our estimate of uncertainty by showing that higher entropy correlates with higher prediction error on the downstream task and enables filtering of low quality generations (i.e. the use of the predicted uncertainties lowers misclassification rate/decision cost). The use of the predictive performance of some downstream discriminative model to evaluate the quality of generative model outputs is a common strategy \cite{kim2021classification,shmelkov2018good}; DTUQ provides a formal justification for this approach in the context of uncertainty quantification.

The DTUQ framework allows us to assess the reliability of our indicators of the trustworthiness of the performance of the generative model (i.e. classification entropy) without the need for ground truths for the generated outputs (instead, we use ground truths of a downstream classifier model). Furthermore, the expression of uncertainty and reliability analysis ties into the ultimate use-case of the adapted outputs, which improves the extent to which our indicators of trustworthiness for the generative model express the practical utility of the adapted examples. This also allows us to use UQ techniques on the downstream classifier instead of the generative model, which is appealing given these are often originally developed for supervised models. The decision-theoretic approach has precedent for large language models \cite{wang2024subjective}. 

\section{Methods}
\begin{figure*}[t]
    \centering
    \includegraphics[width=1.2\columnwidth, trim={0 0 0 1.05cm}, clip]{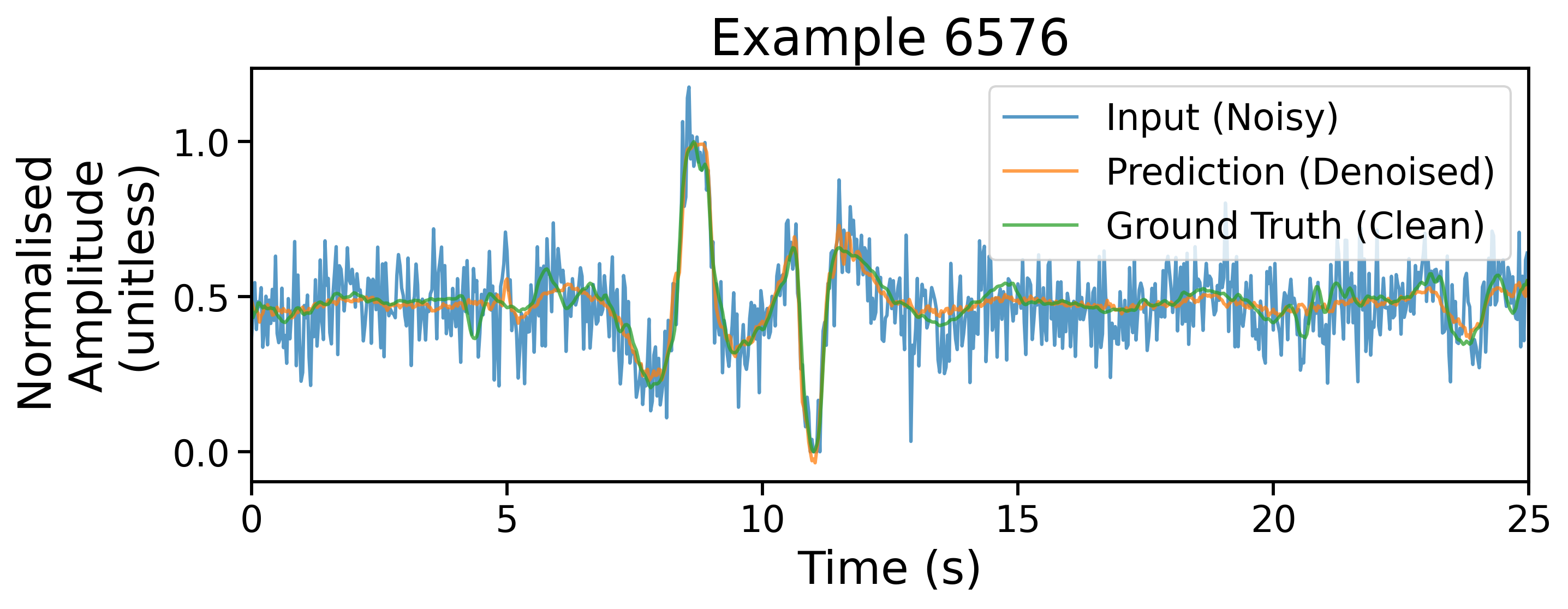}
    \includegraphics[width=1.2\columnwidth, trim={0 0 0 1.05cm}, clip]{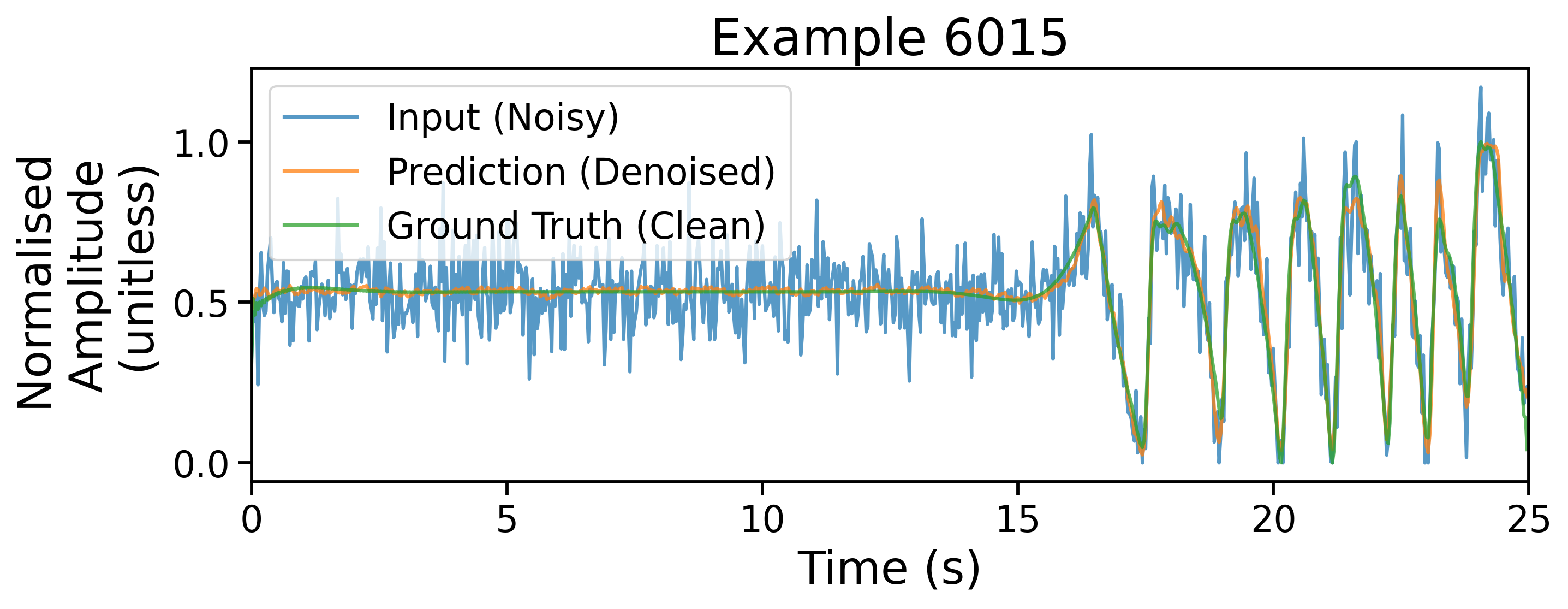}
    \includegraphics[width=1.2\columnwidth, trim={0 0 0 1.05cm}, clip]{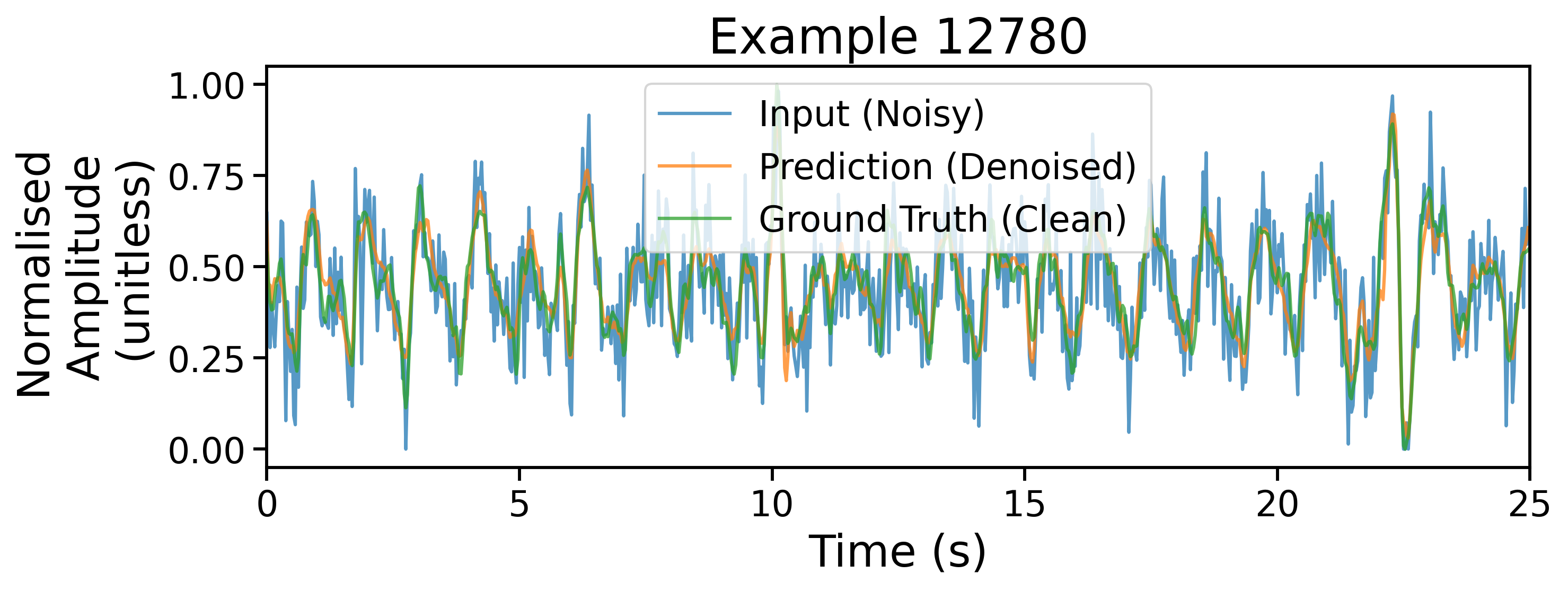}
    \caption{Example PPG times series (noisy, denoised, and the ground truth non-augmented). Top two rows are AF examples. Negative values of the GAN-denoised signals are clamped to zero before being used as inputs to the AF classification model.}
    \label{fig:timeseries}
\end{figure*}

\subsection{Classification model}
Here we consider a 1D variant of AlexNet trained with stochastic gradient descent on a custom split of the Deepbeat dataset as described in \cite{bench2025uncertainty}. We use a custom loss described in \cite{bench2025uncertainty,kendall2017uncertainties}, and evaluation/sampling procedure also given in \cite{bench2025uncertainty}. The model classifies whether an episode of Atrial Fibrillation had occurred within the duration of the time series.

\subsection{Data augmentation}
We add random Gaussian noise (drawn from a distribution with a standard deviation of 0.1) to the entire set of PPG time series, and clamp values to a minimum of 0, and a maximum of 2 (the unaugmented PPG signals are normalised so that global range is $[0,1]$). The noisy version of the training examples are paired with their unaugmented counterparts and used to train the denoising GAN. The same training splits of examples were used for the AF classifier and the GAN.
\subsection{Denoising GAN}
We train a 1D variant of the Pix2pix GAN \cite{henry2021pix2pix}. The model has a UNet backbone \cite{ronneberger2015u} for the generator with 7 encoding layers (1D convolutional layers with kernel size 4, stride of 2, padding of 1, a leaky ReLU (coefficient of 0.2) and 1D instance normalisation on the intermediate blocks). The corresponding 7 decoding layers use 1D transposed convolutions (kernel 4, stride 2, padding of 1) with ReLU activations. Other details can be found in Appendix \ref{sec:model}.

\begin{figure*}[t]
    \centering
    \includegraphics[width=0.6\columnwidth]{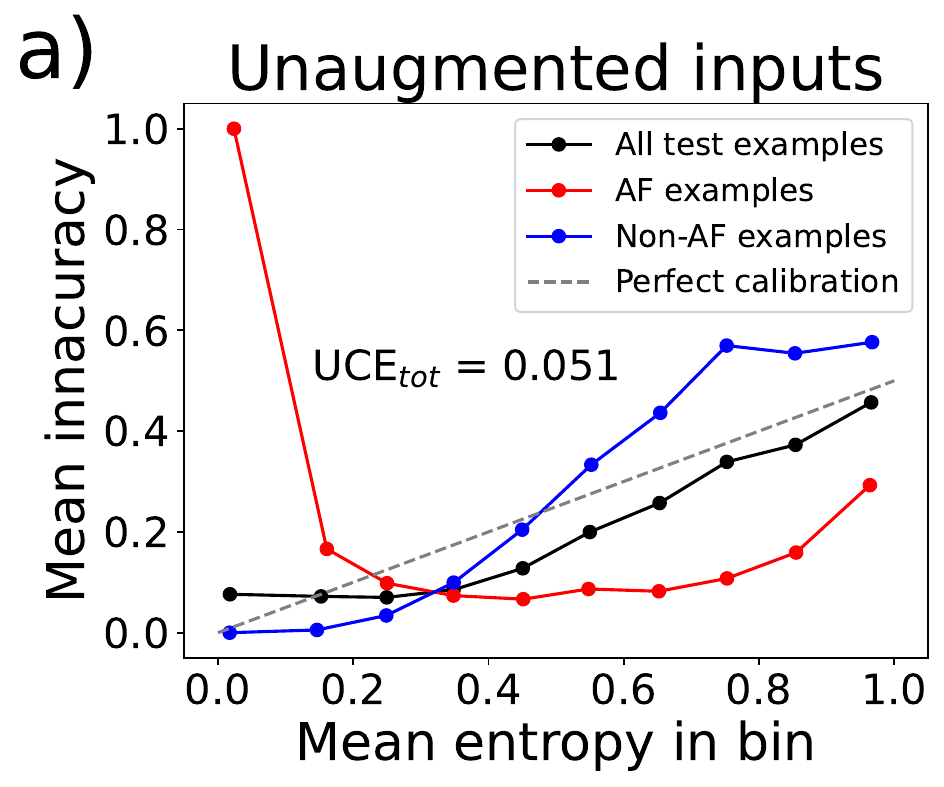}
    \includegraphics[width=0.6\columnwidth]{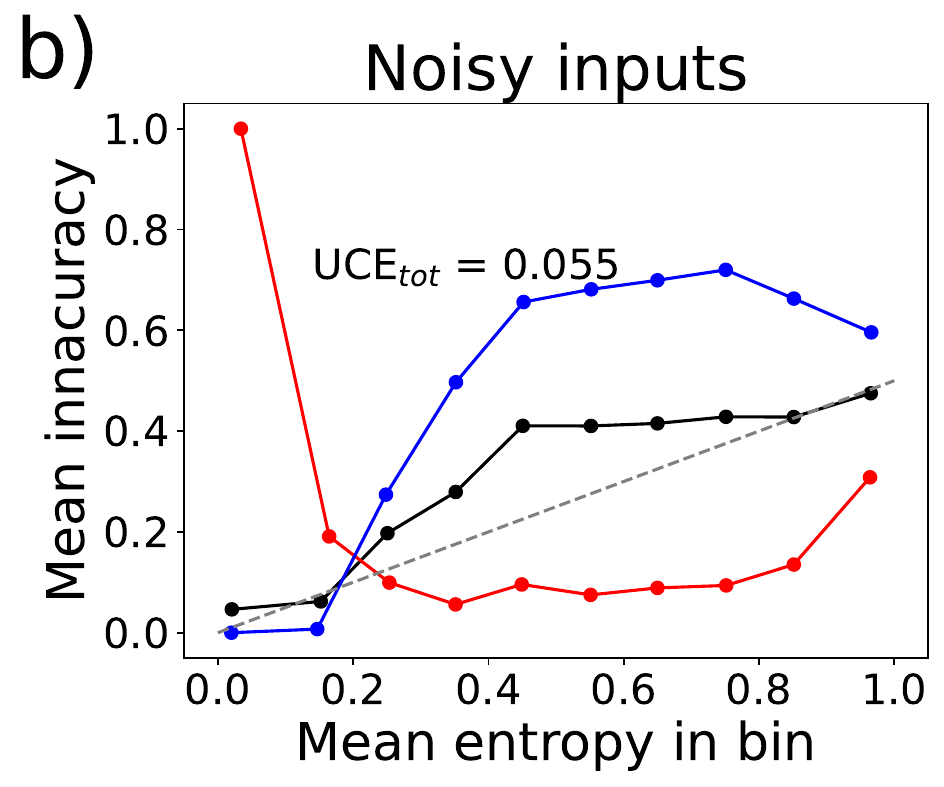}
    \includegraphics[width=0.6\columnwidth]{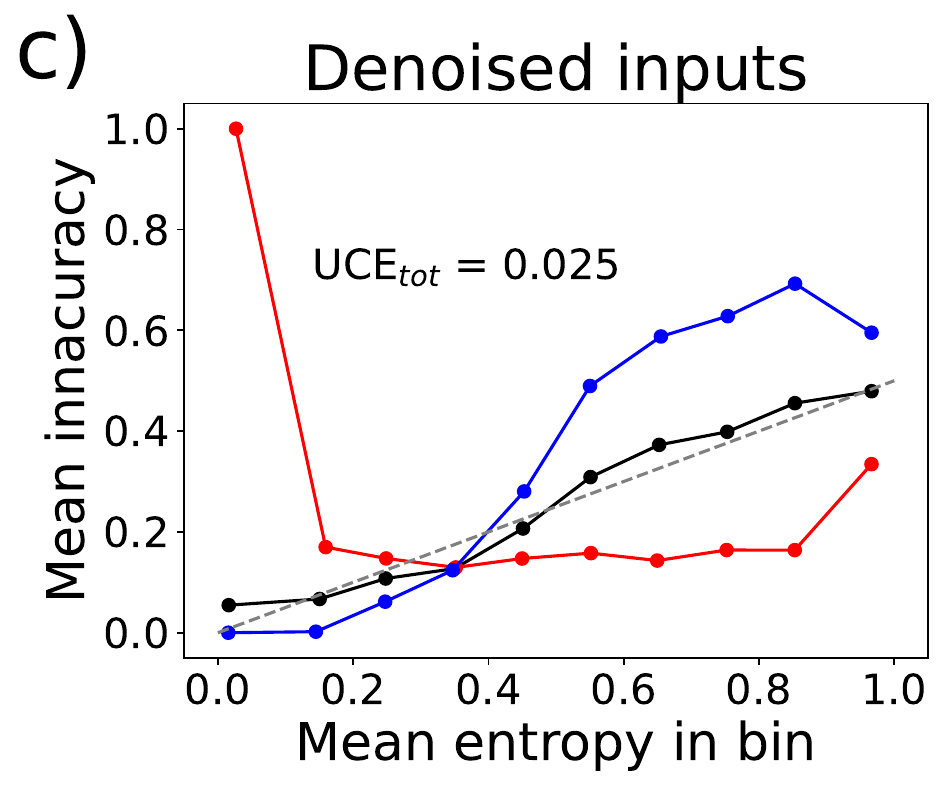}
    \caption{Per-class reliability diagrams for each test set. a) shows the reliability diagram for unaugmented time series, b) shows the same plot for noise augmented time series, while c) shows the uncertainty reliability analysis for the denoising GAN.}
    \label{fig:UCEs}
\end{figure*}

\begin{figure}[h!]
    \centering
    \includegraphics[width=.9\columnwidth]{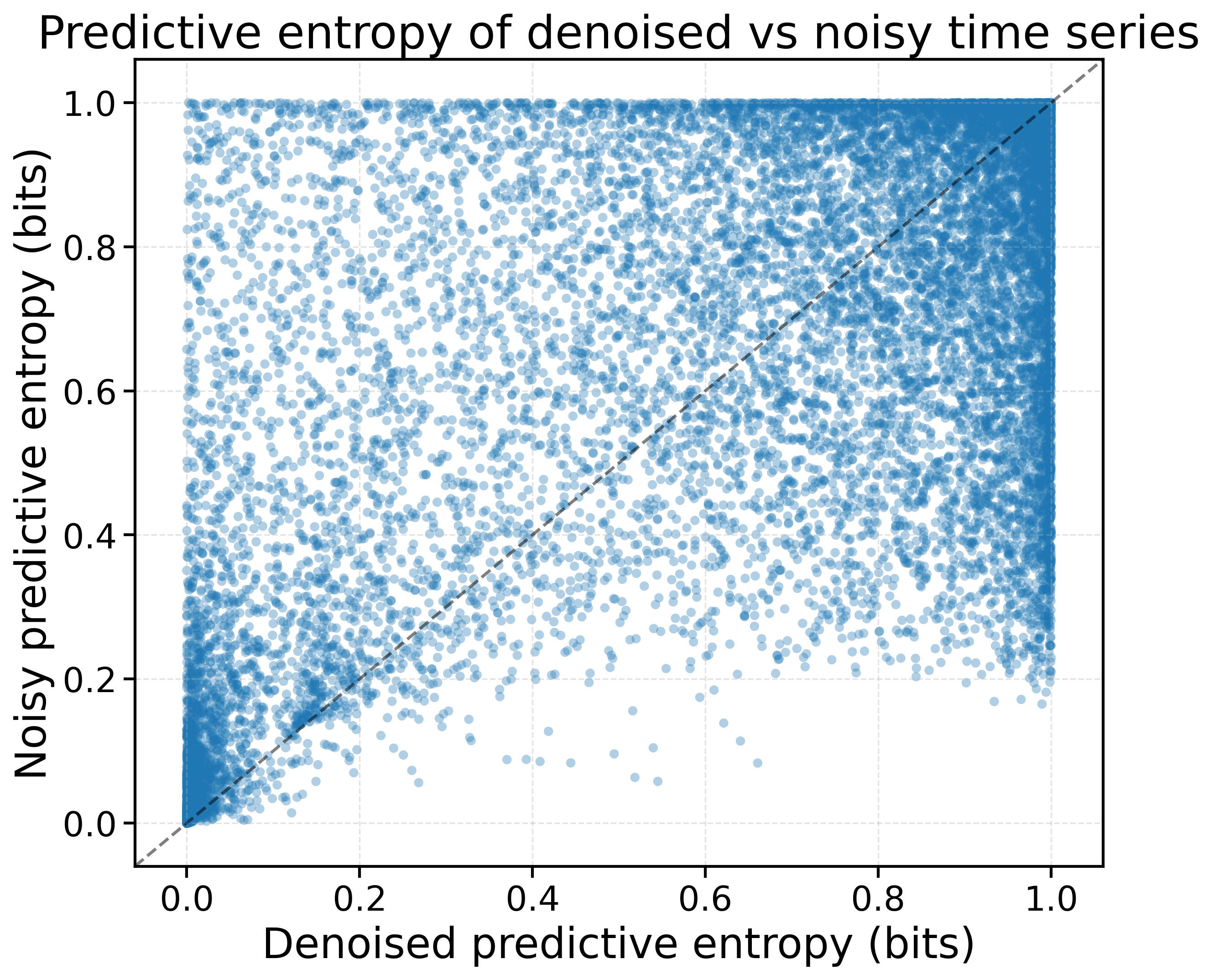}
    \caption{Scatterplot of noisy and denoised predictive entropy. Moderate correlation (as opposed to a strong correlation) suggests that uncertainties reflect changes to time series imposed by the GAN, as opposed to just the underlying properties of the measurement.}
    \label{fig:scatter}
\end{figure}

\begin{table*}[htbp]
\centering
\caption{AF classification performance metrics across conditions. The low uncertainty subset is composed of the denoised examples with corresponding uncertainties in the lower 75 \% of uncertainty magnitudes.}
\label{tab:metrics_snr}
\begin{tabular}{|l|c|c|c|c|c|c|c|}
\hline
Condition & AUC & F1 & \shortstack{ MCC \\(80 \% Sensitivity)} & \shortstack{ MCC \\(80 \% Specificity)} & \shortstack{ Sensitivity \\ (80\% Specificity)} & \shortstack{Specificity \\ (80\% Sensitivity)}& \shortstack{Balanced Accuracy \\ (decision threshold 0.5)} \\
\hline
Unaugmented & 0.84 & 0.71 & 0.51 & 0.50 & 0.71 & 0.72 & 0.76 \\
\hline
Noisy & 0.75 & 0.65 & 0.37 & 0.26 & 0.45 & 0.58 & 0.69 \\
\hline
Denoised & 0.80 & 0.66 & 0.43 & 0.37 & 0.56 & 0.64 & 0.71 \\
\hline
\shortstack[l]{Denoised low-\\uncertainty subset}  & 0.85 & 0.70 & 0.52 & 0.49 & 0.70 & 0.74 & 0.77 \\
\hline
\end{tabular}
\end{table*}

\subsection{Uncertainty quantification}
We express the uncertainty in a denoised time series as the entropy of the predicted distribution of the downstream AF classification model. We assess uncertainty reliability and externally ground uncertainty estimates by observing whether generated outputs with higher uncertainty tend to correlate with lower classification performance. We evaluate this with the Uncertainty Calibration Error and a corresponding reliability diagram (evaluated per class, as well as over the entire test set of examples).

The Uncertainty Calibration Error (UCE) compares the difference between the model's inaccuracy and the average normalised entropy for equal width bins of uncertainty magnitude~\cite{lavesuncertainty}. These differences are weighted by the empirical probability of finding the entropy values within the current bin, $\frac{|B_{m}|}{N}$, where $|B_{m}|$ is the size of bin $m$, and $N$ is the total number of data points, and $k$ is indexes the class. The normalised entropy is given by,
\begin{equation}
    \Tilde{\mathcal{H}}(\mathbf{p}) = -\frac{1}{\log K} \sum_{k=1}^{K} p_{k} \log p_{k}.
\end{equation}

Our heuristic measure of perfect calibration for our binary classifier corresponds to a normalised entropy that is half the misclassification rate; this is evaluated as:

\begin{equation}
    \text{UCE} = \sum_{m=1}^{M} \frac{|B_{m}|}{N} |\text{err}(B_{m}) - \frac{1}{2}\text{uncert}(B_{m})|,
    \label{eq8}
\end{equation}
where $\text{err}(B_{m})$ is the average inaccuracy for all samples in bin $m$, and $\text{uncert}(B_{m})$ is the average normalised entropy for all samples in bin $m$. For the binary classification, a slope of 0.5 for the corresponding reliability diagram indicates perfect calibration. We also consider more generic performance metrics, such as Area Under Operating Curve (AUC), F1 score, Mathews Correlation Coefficient at 80 \% specificity and at 80 \% sensitivity, as well as the specificity at 80 \% sensitivity, the sensitivity at 80 \% specificity and balanced accuracy with a decision threshold of 0.5. While in principle ground truths for the adapted examples are available, we do not use them here to illustrate the capacity of the DTUQ approach to facilitate reliability analysis in cases where these are not available.

\section{Results and discussion}

Table \ref{tab:metrics_snr} shows that adding noise to the time series degrades performance, while using GAN-denoised samples improves predictive performance across all metrics. The low uncertainty subset of denoised predictions consists of predictions where the corresponding entropy was in the lower 75 \% of all predictions. The improvements in performance metrics here suggest that higher uncertainty estimates correspond with generated outputs that are likely to produce inaccurate classifications. This externally grounds the reliability analysis.

The reliability diagrams in Fig. \ref{fig:UCEs} show that the predicted uncertainties exhibit some degree of reliability, where there is significant discrepancy in the extent of the calibration for each class. For all datasets considered, all of the predictions in the lowest entropy bin for AF examples are misclassified as non-AF, suggesting poorer reliability for this class (and for the subset of examples with very low entropy). These results align with those reported in related work considering the same models/datasets and optimisation scheme \cite{bench2025uncertainty}. Generally, denoised samples that have high predictive entropy are more likely to be classified incorrectly. Assuming this calibration generalises, an uncertainty estimate could be used to improve the use of the denoising GAN/filter out examples that are likely to produce incorrect classifications.

It is possible that the uncertainty estimation is not sensitive to the features/artefacts imposed on the time series by the GAN, and instead, is sensitive to the underlying properties of the measurement. We provide a scatterplot (Fig. \ref{fig:scatter}) comparing entropy of the predictions for the noisy and denoised versions of each example to show the extent to which denoising can change the estimate of uncertainty, and hence, whether the uncertainty reflects significant changes to the time series imposed by the GAN. We calculate a Pearson's correlation coefficient of 0.68, and a Spearman's rank correlation of 0.59. Moderate correlation (as opposed to a strong correlation) suggests that the uncertainty reflects significant changes to the time series imposed by the GAN, and may be used to filter out poor quality generations rather than just poor quality measurements.

Given the aim here is to improve classification performance on a downstream classification model, decision cost is trivial to define (i.e. misclassification loss). However, defining decision cost in the case where we do not rely on a downstream classifier is more challenging. This should be carefully formulated based on the user's preferences and external grounding by comparing whether higher uncertainties result in lower decision cost. In principle, other performance metrics for the downstream classifier (e.g. precision or recall) could be considered instead of misclassification loss, where some metrics may have less trivial expressions of the Bayes risk (e.g. F1 score as it relies on global assessment of true/false positives and false negatives, which makes it challenging to formulate a per-instance indicator of trustworthiness).

\section{Conclusion}
We have described how the use of the predictive entropy of a  downstream classification model to assess the quality of data produced from a generative model can be formalised as an implementation of decision-theoretic uncertainty quantification. This enables an assessment of the reliability of our chosen indicator of the trustworthiness of generated outputs through the use of the Uncertainty Calibration Error, and facilitates trustworthiness assessment for deep domain adaptation in cases where the adaptation is used to improve the performance of a downstream classifier. This helps demonstrate the utility of the DTUQ framework for generative deep domain adaptation in PPG analysis and otherwise.

\appendix
\subsection{GAN implementation}
\label{sec:model}
Long range skip connections are included in the generator to concatenate encoder activations with decoder features at matching scales. Normalisation is given by 1D instance normalisation with  affine set to false, except in the outermost block. The final layer uses a tanh activation. The discriminator is a fully convolutional 1‑D PatchGAN \cite{li2016precomputed} that outputs a 1‑channel logit map along time. Stacked Conv1d blocks (kernel 4, stride 2, padding of 1) with LeakyReLU progressively downsample the sequence to produce patch‑level classifications. Instance normalisation is used in all but the first block. A learning rate of 1e-5 was used for the discriminator, and 2e-4 for the generator. The discriminator loss is given by,

\begin{align}
\mathcal{L}_D
&= 
\frac{1}{2} \left(
\| D(A, B) - 1 \|_2^2
+
\| D(A, G(A)) \|_2^2
\right),
\end{align}
and the generator loss $\mathcal{L}_G$ is given by,
\begin{align}
\mathcal{L}_G
&=
\mathcal{L}_{G,\text{GAN}} + \mathcal{L}_{G,\ell_1},
\end{align}
where,
\begin{align}
\mathcal{L}_{G,\text{GAN}}
&=
\| D(A, G(A)) - 1 \|_2^2, \\[6pt]
\mathcal{L}_{G,\ell_1}
&=
\lambda_{\ell_1} \, \| B - G(A) \|_1, \\[6pt]
.
\end{align}
where $A$ is a noisy signal, and $B$ is a target non-augmented signal, $D$ is the discriminator and $G$ is the generator and $\lambda_{\ell_1}=100$. The subtraction of 1 here is performed across all elements of the discriminator's output. The discriminator is updated first then the generator for each batch. Early stopping on the validation set $\mathcal{L}_{G,\ell_1}$ loss with a patience of three epochs was implemented. The discriminator takes a real pair (a real noisy time series and its unaugmented counterpart) and a fake pair of examples (a real noisy time series and its generator denoised counterpart) as inputs. The generator takes noisy time series as inputs.
\bibliographystyle{ieeetr}
\bibliography{references.bib}

\end{document}